\documentclass[sigplan]{acmart}
\usepackage{gensymb}
\usepackage{soul}
\usepackage{graphicx}
\usepackage{amsmath}
\usepackage{amssymb}
\usepackage{multirow}
\usepackage{subfigure}
\usepackage[none]{hyphenat}
\usepackage{epsfig}
\usepackage{epstopdf}

\emergencystretch=\maxdimen
\setcopyright{none}
\renewcommand\footnotetextcopyrightpermission[1]{}
\begin{document}
\title{On the Inference of Soft Biometrics from Typing Patterns Collected in a Multi-device Environment}

\author{Vishaal Udandarao}
\email{vishaal16119@iiitd.ac.in}
\authornotemark[1]
\affiliation{%
  \institution{IIIT Delhi, India}}

\author{Mohit Agrawal}
\email{mohit.nittrichy@gmail.com}
\affiliation{%
  \institution{IIIT Delhi, India}}
\authornote{Both authors contributed equally to this research.}

\author{Rajesh Kumar}
\email{rkumar@haverford.edu}
\affiliation{%
  \institution{Haverford College, USA}
  \city{Syracuse University, USA}}

\author{Rajiv Ratn Shah}
\email{rajivratn@iiitd.ac.in}
\affiliation{%
  \institution{IIIT Delhi, India}
}

\renewcommand{\shortauthors}{Udandarao and Agrawal, et al.}

\begin{abstract}
\quad In this paper, we study the inference of gender, major/minor (computer science, non-computer science), typing style, age, and height from the typing patterns collected from 117 individuals in a multi-device environment. The inference of the first three identifiers was considered as classification tasks, while the rest as regression tasks. For classification tasks, we benchmark the performance of six classical machine learning (ML) and four deep learning (DL) classifiers. On the other hand, for regression tasks, we evaluated three ML and four DL-based regressors. The overall experiment consisted of two text-entry (free and fixed) and four device (Desktop, Tablet, Phone, and Combined) configurations. The best arrangements achieved accuracies of 96.15\%, 93.02\%, and 87.80\% for typing style, gender, and major/minor, respectively, and mean absolute errors of 1.77 years and 2.65 inches for age and height, respectively. The results are promising considering the variety of application scenarios that we have listed in this work. \end{abstract}

\ccsdesc[500]{Security and privacy~Human and societal aspects of security and privacy}

\keywords{Privacy, soft biometrics, keystroke dynamics, typing patterns, gender recognition, multi-device, applied machine and deep learning}



\maketitle

\section{Introduction}
"Everyone is special, and nobody is like anyone else. Everyone's got an act."-- The
Greatest Showman. \\

While we interact with computing devices, we leave a variety of footprints such as typing, swiping, walking, among others. These footprints have been studied for authentication, identification, forensic analysis, health monitoring, cognitive assessment, and inferring soft biometric traits  \cite{KeystrokeSurveyDamaon2012,KeyStrokeHealthMonitoring2015, SoftBiomOnMobile2019,AgeGenderHandMobile2016,TypingCognition,WhatElseRoss2016,SoftBiometricsNixon2015,SwipeToGender2016}. Typing \cite{KeystrokeSurveyDamaon2012,KeystrokeSurvey,KeyStrokeSound,KeystrokeVideo}, swiping \cite{Touchalytics,patel2016continuous,TowardRobotic2016}, gait \cite{kumar2018continuous,kumar2016authenticating,kumar2015treadmill,primo2014context}, body movements \cite{kumar2017continuous}, and fusion are some of the widely studied behavioral patterns in the context of desktop, mobile, and wearable devices. Typing is commonly characterized as key press and release timings, keystroke sounds, and video sequence \cite{KeystrokeSurveyDamaon2012,KeystrokeSurvey,KeyStrokeSound,KeystrokeVideo}. Security critical organizations such as the Defense Advanced Research Projects Agency (DARPA) have already adapted typing-based active authentication technology for desktops \cite{DARPAAA}. 

However, the majority of the keystroke studies focus on either authentication or identification under free or fixed-text entry environments \cite{KeystrokeTimingMovement2016, KeystrokeSurvey, KeystrokeSurveyDamaon2012, belman2020discriminative}. The number of studies on the inference of soft biometrics from typing patterns is limited or confined to a particular device/environment or both \cite{FirstGenderFromKeystroke2011,ImprovedKDAuthWithGenderPred2012,plank2018predicting,Gender2018FromKeystroke,CombiningKDandStylometryGenderPredFromChat2019,GenderFromChat2020,AgeGendernumFingers2014,GenderAgeKeystrokeHamdedness2014FreeFixedBoth,AgeGenderHandMobile2016,CombineKDandSignatueForGenderPrediction,ChildVsAdultKeystroke,AgeGenderKDMouseMove2017}. Inference of a variety of personal attributes including but not limited to age, gender, cognitive assessment, handedness, typing hand, and number of fingers used for typing have been explored in the past \cite{Gender2018FromKeystroke,GenderRecognitionFroMKeystrokeAndSwipes,AgeGenderKDMouseMove2017,GenderAgeKeystrokeHamdedness2014FreeFixedBoth, AgeGenderHandMobile2016,SoftBiometricsIsLessExplored}. Considering that typing is an indispensable part of our lives, we believe that it reveals a great deal of information and should be studied in depth for the inference of useful identifiers. The identifiers inferred from typing patterns can be used in a variety of ways. Some of them are listed below: 
\begin{figure*}[htp]
\centering
\includegraphics[width=6.8in, height=1.45in]{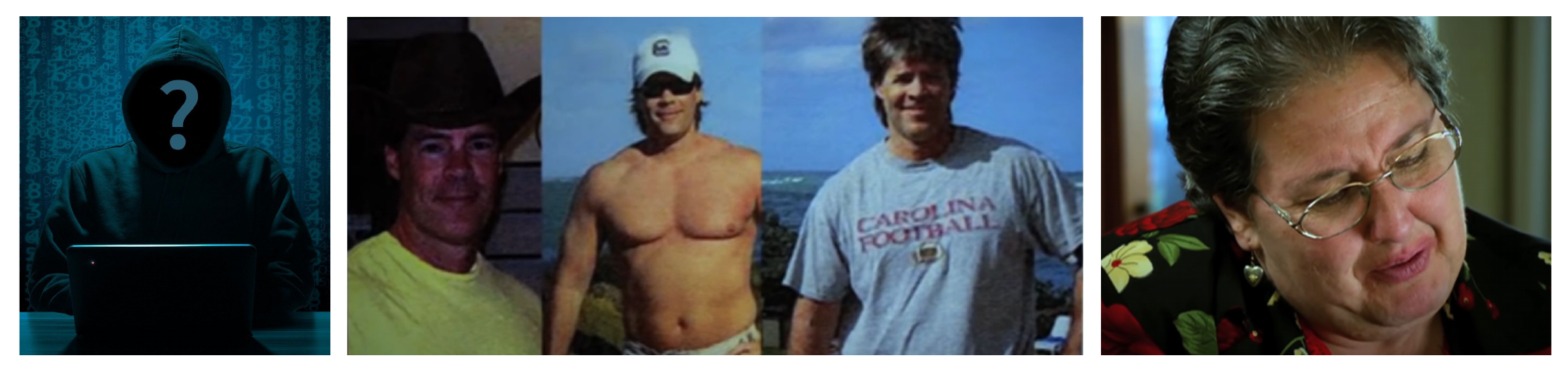}
 \caption{Person (on the left) impersonated Benjamin (in the middle, a handsome American businessman) to fool a divorced and lonely woman Rosely (on the right) and scam out her lifelong savings ($\$90,000$) by promising her lifelong love \cite{LoveScamAustralia}. The typing patterns of the person could have been used to estimate the gender, age, height, and weight, and alarm Rosely that the person she is thinking the love of her life may be fake as his/her soft traits do not match with the information provided to her. Besides, law enforcement personnel can use soft biometrics for tracing and convicting the person.}
 \label{LoveScamNigeria}
\end{figure*}
\begin{itemize} 
    \item \textbf{Personalized user experience:} Consumers often refrain from providing too much information while signing up for an information technology-enabled service. Besides, people with disabilities may find it difficult to enter too much information to start using a software platform. The automated estimation of soft biometrics can be useful in such cases. Organizations can tailor their platforms and services as per the user's demography for a seamless and personalized experience. Moreover, estimated soft biometrics can be used for controlling access to certain resources or platforms. For example, access to certain TV channels and websites can be restricted to individuals of certain age groups. 
    
    \item \textbf{Improved recognition rate:} The performance of an authentication and identification systems can be improved by incorporating the inferred soft biometrics such as age, gender, weight, and height in the pipeline \cite{SoftBiometricsIsLessExplored, WhatElseRoss2016, SoftBiomeEnhanceKDRecognition2019, GenderAgeKeystrokeHamdedness2015}.
    
    \item \textbf{Targeted advertising:} Organizations can use the soft biometrics for customized their advertisement and target people of a specific height, weight, gender, and age groups who might be interested in the product more than the rest \cite{SoftBiometricsIsLessExplored, WhatElseRoss2016}. 
    
    \item \textbf{Identification of fake profiles on social media:} The social-media platforms are suffering from fake profiles and fake news spread. It is not uncommon for individuals to fake their identity, i.e., to be a different gender, height, age, and profession. It is difficult to determine the legitimacy of individuals based on the type of information they post. The accurately estimated soft identifiers based on the typing pattern can help detect these profiles and take appropriate actions \cite{CombiningKDandStylometryGenderPredFromChat2019, FirstGenderFromKeystroke2011}.
    
    \item \textbf{Forensics:} Covert identification of individuals has never been more critical than today as the number and nature of cybercrimes are rapidly evolving \cite{CombiningKDandStylometryGenderPredFromChat2019}. As per Federal Bureau Investigation (FBI)'s 2019 Internet Crime Report, $467,361$ online scams were registered alone in 2019 \cite{FBI2019CrimeReport}. These scams cost innocent people a total of $\$3.5$ billion. Business email compromise, romance fraud, and spoofing caused the highest financial losses. Several victims ended up losing their entire life savings or even sinking into debt. The law enforcement agencies often lack credible information to trace and convict these scammers. Soft biometrics inferred from typing footprints that the scammers leave while they interact with the victims could be useful in such scenarios (see Figure \ref{LoveScamNigeria} for an example). 
\end{itemize}

The above-mentioned applications motivated us to study the inference of soft biometrics from typing patterns of individuals in a multi-device environment. In summary, this work makes the following set of contributions:
\begin{itemize}
    \item Investigate inference of five soft biometrics, namely, gender, major/minor, and typing style, age, and height from typing patterns collected from 117 individuals while they typed a predefined text and answered a series of questions on a desktop, tablet, and smartphone.
    
    \item Benchmark six Machine Learning (ML) and four Deep Learning (DL) algorithms for the classification of gender, major/minor, and typing style. Additionally, we benchmark eight different configurations generated from two factors (free and fixed-text entry), and devices (Desktop, Phone, Tablet, and Combined). 
    
    \item Besides using unigraphs, digraphs, and word-level features with a mutual information-based feature selector, we explore a novel method of constructing the feature space for the application of DL methods.

    \item Provide detailed results and discussion on the inference of gender, major/minor, typing style, age, and height of the participants. Besides, present a qualitative performance comparison with the existing studies.
    
    \item Share the code base for reproducibility of results and foster future research in this direction.\footnote{Code is available upon request. Please send an email to the last author.} 

\end{itemize}

The rest of the paper is organized as follows. Section \ref{RelatedWorks} discusses the closely related works. Section \ref{DesignOfExperiments} presents the design of experiments. Section \ref{sec:Results}, and Section \ref{sec:Discussion} present and discuss the results, respectively. Finally, we conclude the paper and provide future research directions in Section \ref{sec:ConclusionAndFutureWork}. 

\section{Related work}
\label{RelatedWorks}
The inference of soft biometrics (gender, age, ethnicity, hair/eye/skin colors, and hairstyle) from physical biometrics (e.g., face, fingerprint, iris, hand, and body), as well as gait and voice, have been substantially covered by Dantcheva et al. \cite{WhatElseRoss2016}. Thus, in this section, we describe the works related to the inference of soft biometric from typing patterns, and the gap that this work attempts to fill in.

Early attempts to infer the gender of the typists from keystroke analysis were made in \cite{FirstGenderFromKeystroke2011, ImprovedKDAuthWithGenderPred2012}. One \cite{FirstGenderFromKeystroke2011} was inspired by developing trust and reliability among social network users, while the other \cite{ImprovedKDAuthWithGenderPred2012} was motivated from improvement in the performance of user recognition systems by including estimated soft-biometrics as features. For example, Idrus et al. \cite{ImprovedKDAuthWithGenderPred2012, GenderAgeKeystrokeHamdedness2015} utilized the determined gender, age, and handedness to achieve about 7\% of reduction in user recognition error rate. A separate study by Idrus et al. \cite{GenderAgeKeystrokeHamdedness2014FreeFixedBoth} was conducted under fixed- and free-text entry environment to predict the hand category (use one or both hands), gender (male, female), age ($<30$ or $\ge 30$), and dominant hand (lefty or righty). Brizan et al. \cite{TypingCognition} used hybrid (keystroke, stylometry, and language production) set of features to predict the cognitive demands of a given task. Yasin et al. \cite{ChildVsAdultKeystroke} were able to differentiate between children (below 15) and adults (above 15) by analyzing the participant's typing behaviors. Recently, Abeer et al. \cite{GenderFromChat2020} predicted gender from live chats. 

Pentel \cite{AgeGenderKDMouseMove2017} combined mouse patterns with keystrokes to predict the age and gender of individuals. Likewise, Li et al. \cite{CombiningKDandStylometryGenderPredFromChat2019} analyzed stylometry and keystroke dynamics to predict the gender of the person from 15 minutes of chat with 72\% accuracy. Bandeira et al. \cite{CombineKDandSignatueForGenderPrediction} combined handwritten
signature and keystroke dynamics for gender prediction. Abreu et al. \cite{ThreeModalGenderRecognition2019} also combined three modalities (keystrokes, touch strokes, and handwritten signature) to predict the gender of the typists. The authors suggested that the fusion-based system outperformed the rest. Buriro et al. \cite{AgeGenderHandMobile2016} estimated age, gender, and operating hands from the typing behavior of individuals collected on smartphones. 

Other than age, gender, handedness, and dominant hand, researchers have predicted some interesting indicators from typing patterns. For example, Epp et al. \cite{EmotionfromKeyStrokes2011} investigated the prediction of fifteen emotional states, including confidence, hesitance, nervousness, relaxation, sadness, and tiredness from typing patterns. Tsimperidis et al. \cite{KeystrokeDynamicsBasedEducationalLevel2020} predicted the educational level of participants based on the keystroke dynamics information only. Beyond typing patterns, researchers have explored other behavioral patterns such as walking patterns, swiping patterns, calling patterns, device usage patterns to estimate a wide variety of soft identifiers \cite{MobileCallingPattern, TouchToGender2016, SurveyOnGaitToAgeAndGender2019, TouchToAge2018, MobileDeviceUsageGender2018}.

The aforementioned studies have shown that an individual's behavioral pattern reveal about their gender, age, handedness, dominant hand, emotional stress, cognitive ability, etc. These studies, however, were limited in terms of types of devices used in the experiments, data collection protocol (free or fixed text), application of algorithms, and prediction of specific soft biometric. The majority of the application scenario mentioned in the introduction would require the study on the inference of soft biometrics from behavioral patterns to be more thorough. By thorough, we mean the inclusion of a variety of users, devices, text entry mode, and a variety of learning paradigms that could be more suitable, in addition to collecting the absolute ground truth. 

Conducting such a comprehensive study on this topic would require a grand data collection experiment. One of the datasets that aligned well with our hypothesis is the dataset recently posted by Belman et al. \cite{AmitBBMAS2019}, which includes fixed as well as free text collected from 117 users who answered a wide variety of questions on a desktop, tablet, and smartphone. The specific soft traits that we included in this study are age, gender, height, typing style (must look at the keyboard, occasionally looks at the keyboard, and need not look at the keyboard), major/minor (computer science or non-computer science). Apart from considering five soft traits, we study keystroke features that (e.g., word-level features) have not been studied in this context but shown to be better than traditional keystroke features in the context of user recognition \cite{WordLevelFeaturesAreBetter,belman2020discriminative}. Moreover, we apply numerous learning algorithms, which have not been studied in this context before, to the best of our knowledge.


\section{Design of experiments}
\label{DesignOfExperiments}
\subsection{Dataset}
We used Syracuse University and Assured Information Security-Behavioral Biometrics Multi-Device and Multi-Activity Data from the Same Users (SU-AIS BB-MAS) \cite{AmitBBMAS2019}. The dataset consists of multiple modalities; however, we consider only the keystroke part, therefore refer to the dataset as BB-MAS-Keystroke in this document. 

The BB-MAS-Keystroke consists of $3.5$ million keystrokes collected from $117$ users who typed two given sentences (fixed) and answered a series of questions (free-text) on desktop (Dell kb212-b), tablet (Samsung-S6), and smartphone (HTC-Nexus-9). A summary of the dataset is provided in Table \ref{tab:samplestat}. Please see \cite{AmitBBMAS2019} for more details. 

\subsection{Feature extraction and analysis}
\begin{figure}[htp]
    \centering
    \includegraphics[height=2in, width=3.5in]{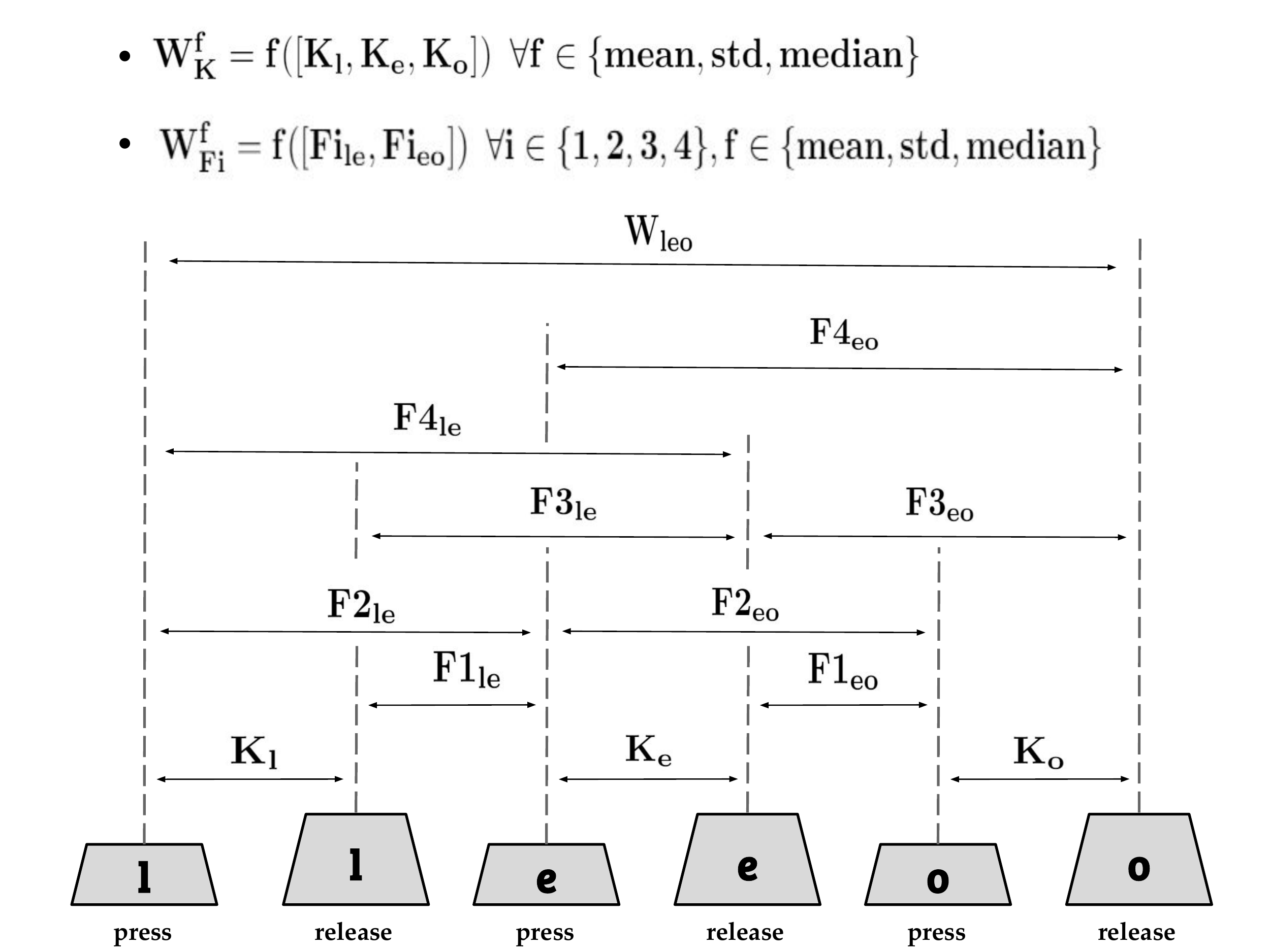}
    \caption{An illustration of the extraction of unigraphs $(K)$, digraphs $(F)$, and world-level $(W)$ features.}
    \label{fig:features}
\end{figure}

Following previous studies \cite{WordLevelFeaturesAreBetter,belman2020discriminative, huang2016effects, KeystrokeSurvey}, we extracted unigraph (Key Hold Time), digraph (Flight or Key Interval Time), and word-level features. Before feature extraction, we removed outlier using interquartile range (IQR) method. The description of features computation is provided below and pictured in \autoref{fig:features}:
\begin{itemize}
    \item \textbf{Unigraphs:} Unigraphs are defined as the difference between the key release and key press timings. These features were extracted for all unigraphs in the data and aggregated. For example, if the key $k$ is pressed and released $50$ times in the dataset, the key hold feature of $k$ would be a list of $50$ values.
    \item \textbf{Digraphs:} Digraph captures information about the press and release timings of two consecutive keys. There are four different digraphs that can be defined for two consecutive keys (say $k_i$ and $k_{i+1}$) as demonstrated as follows: 
    \begin{enumerate}
        \item \textit{$F1 = (k_{i+1})_{press}-(k_{i})_{release}$}
        \item \textit{$F2 = (k_{i+1})_{release}-(k_{i})_{release}$}
        \item \textit{$F3 = (k_{i+1})_{press}-(k_{i})_{press}$}
        \item \textit{$F4 = (k_{i+1})_{release}-(k_{i})_{press}$}
    \end{enumerate}
    We observed that in some cases, the key $k_{i+1}$ was pressed before the release of key $k_i$, which resulted into negative values for the features $F1$ and $F3$ for those occurrences. The aggregation process was same as unigraphs.
    \item \textbf{Word level features}: The word-level features capture different characteristics of the data than the uni and digraphs. They are also shown to be highly discriminative among users \cite{WordLevelFeaturesAreBetter,belman2020discriminative}. Thus, we adapted these features in this study. These features were computed as described as follows: \\
    Consider a word $W$ of length $n$ consisting of the keys \{$k_1$, $k_2$, ..., $k_n$\} in that order. Then word-level features were defined and extracted as follows: 
    \begin{enumerate}
        \item \textbf{Word Hold Time ($W_N = (k_{n})_{release}-(k_{i})_{press}$}
        \item \textbf{Word-unigraph features ($W_K^{f}$)}: These features consisted of mean, standard deviation, and median of the unigraphs of $W$. Assume we use an aggregation function $f$, then for the word $W$, 
        \[
        W_K^{f} = f([K_{k_1}, K_{k_2}, ..., K_{k_n}])\]
        \item \textbf{Word-digraph features ($W_{Fi}^{f}$)}: Similar to word-unigrah features, we computed the word-digraph features. Assume the aggregation function $f$ and flight features $F_i$ (where $i\in \{1,2,3,4\}$), then for the word $W$,
        \[W_{Fi}^{f} = f([{Fi}_{k_1, k_2}, {Fi}_{k_2, k_3}, ..., {Fi}_{k_{n-1}, k_n}])\]
    \end{enumerate}
\end{itemize}

More details on how these features were utilized during the classification is provided in Section \ref{classicalML} and \ref{DeepLearning}.

\begin{figure}[htp]
    \centering
    \includegraphics[height=0.9in, width=3in]{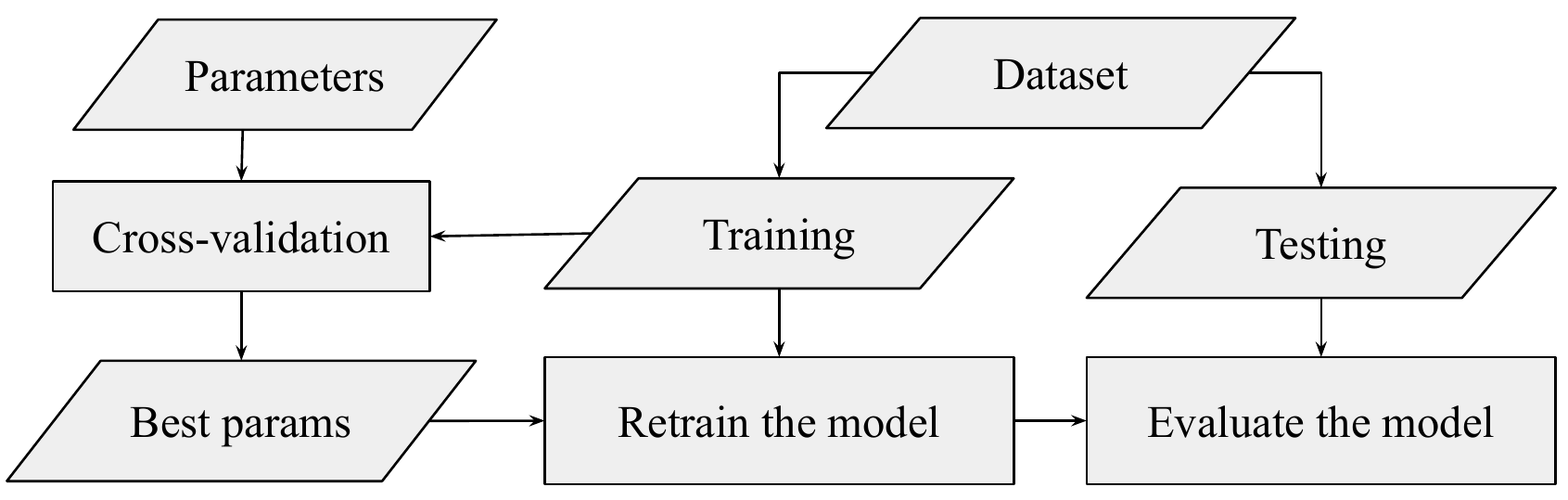}
    \caption{Training, cross-validation, and testing setup. The data was divided in user sets $P$ (Training and cross-validation for hyper-parameter tuning) and $Q$ (Testing). Where, $P \cap Q = \phi$. Adopted from \cite{ScikitCrossVal}.}
    \label{fig:train-test}
\end{figure}

\begin{table}[htp]
\small
\centering
\caption{Number of samples available in the dataset \cite{AmitBBMAS2019}. We studied only the first five as the last two were extremely imbalanced which is one of the limitations of the dataset.}
\vspace{0.03in}
\label{tab:samplestat}
\begin{tabular}{l|l}
\toprule
\multicolumn{1}{c|}{\textbf{Soft biometric}} & \multicolumn{1}{c}{\textbf{Description}}  \\ 
\midrule 
Gender                                      & male (72), female (45)                                                                                                                                                   \\ 
\midrule 

Major/minor                                       & CS(66), non-CS (50), missing (1)                                                                                                                                          \\ 
\midrule 

Typing style                                & \begin{tabular}[c]{@{}l@{}}a: must look at the keypad (6), \\ b: occasional look at the keypad (31), \\ c: need not look at the keyboard (80)\\ \end{tabular} \\ 
\midrule 

Age (years)                                 & \begin{tabular}[c]{@{}l@{}}range (19, 35), mean = 24.97, \\ median = 24.0, std = 3.11\end{tabular}                 
\\  
\midrule 

Height (inches)                             & \begin{tabular}[c]{@{}l@{}}range(54, 74), mean = 66.96,  \\ median = 67.0, std = 4.02\end{tabular}                                                                       \\  
\midrule 

Ethnicity                                   & Asian (104), non-Asian (13)                                                                                                                                              \\  
\midrule 

Handedness                                  & \begin{tabular}[c]{@{}l@{}}right (114), left (1),  \\ ambidextrous (2)\end{tabular}                                                                                      \\ 
\bottomrule
\end{tabular}
\end{table}

\subsection{Learning framework}
Prediction of gender (female or male), typing style (must look at the keyboard or occasionally looks at the keyboard or need not look at the keyboard), and major/minor (computer science or non-computer science) were considered as classification tasks. On the other hand, age and height estimation was considered as regression tasks in our experiments. The block diagram of the learning framework adopted in this study is illustrated in Figure \ref{fig:train-test}. We divided the dataset in two parts \textit{Training} and \textit{Testing}. The \textit{Training} data consisted of 70\% of the users, and as the name indicated, it was used to train the model and tune the hyperparameters using five-fold cross-validation. The best-performing values of the hyperparameters were then used to train the model again on the \textit{Training} dataset. The trained model was then tested on the \textit{Testing} dataset, which consisted of the remaining 30\% users. The adopted learning framework creates a realistic experimental setup as it allowed us to test our model on completely unseen data, unlike some previous works \cite{SwipeToGender2016, Gender2018FromKeystroke, FirstGenderFromKeystroke2011, plank2018predicting, belman2020discriminative}, which have reported the results using k-fold cross-validation on the whole dataset. Nevertheless, we tried this strategy as well and got near-perfect results. 

Also, we observed that the dataset has a class imbalance problem. For example, the number of males was higher than the number of females (see Table \ref{tab:samplestat} for more details). Borderline over-sampling based on SMOTE (Synthetic Minority Oversampling Technique) \cite{nguyen2011borderline} was included in the classification pipeline to over-sample the minority class samples and make it equal to the majority class samples. Borderline SMOTE was chosen over vanilla SMOTE \cite{chawla2002smote} and Adaptive Synthetic (ADASYN) sampling technique \cite{he2008adasyn} based on the loss obtained during training.

\subsubsection{Classical Machine Learning (ML)}
\label{classicalML}
We included a variety of algorithms for implementing the classification and regression tasks. The decision to include algorithms such as Naive Bayes, Decision Trees, Support Vector Machine (SVM), Adaptive Boosting (AdaBoost), and Multi-Layer Perceptron (MLP) with single hidden layer was based on the previous studies
\cite{TypingCognition, AgeGenderHandMobile2016, SwipeToGender2016, morales2016keystroke, baluja2007boosting, neal2018gender, plank2018predicting, KeystrokeForensics2013, GenderIdFroMEmails2009}. Besides, we included algorithms, namely extreme gradient boosting (XGBoost), that have been rarely studied in this context but drew attention due to its success in online competition platforms such as Kaggle \cite{Kaggle}. The hyperparameters associated with these algorithms were tuned using five-fold cross-validation and grid search (see Figure \ref{fig:train-test}).  

In addition to tuning the listed parameter, we also experimented with the number of features and presented the best results obtained. The encouraging performance of ML algorithms, as well as the size of data, motivated us to experiment with deep learning methods that have been effectively used for solving typing pattern-based identification and authentication, recently \cite{dl2, dl3, dl1, dl4}. 

\subsubsection{Deep Learning (DL)}
\label{DeepLearning}
Deep learning has been used with great success in recent years. The combination of deep networks, along with the non-linear activation, has been influential in the popularity of deep learning algorithms. Recently, there have been several attempts at using deep learning architectures for analyzing keystroke biometric data \cite{dl2, dl3, dl1, dl4}. Inspired by these approaches, we leverage the following deep learning models:
\begin{itemize} 
    \item \textbf{Fully Connected (FC) Network}: We use a four-layered neural network with relu activation. We additionally incorporate dropout as a regularization technique for our model. We believe that using a deep FC network will help capture the intrinsic differentiating factors within the aggregated feature vectors to help discern the privacy factors better.
    
    \item \textbf{Convolution Neural Network (CNN)}: We use a seven-layer CNN with four 2D convolution layers and three fully connected layers. We further use dropout and batch normalization to regularize our network. Since our data features are in the form of vectorized arrays, we use a trick of converting them into squared images. For a given feature vector of dimensionality $N$, we find the largest perfect square S just smaller than N and convert the feature vector to an image of size $1 \times \sqrt{S} \times \sqrt{S}$. We hypothesize that the trick will help us leverage CNNs to exploit the structural and spatial biases present in our feature data efficiently.  
    
    \item \textbf{Recurrent Neural Network (RNN)}: We use a three-layer RNN with tanh activation functions and a final softmax classification layer. In the case of RNNs, we require our input data to be sequential. However, our data is in the form of tabulated feature vectors. We use a heuristic to convert our feature vectors into sequential data points to feed it into the RNN. For a given feature vector of dimensionality $N$, we find the largest non-prime number just smaller than $N$ and find two factors $A$ and $B$ such that $N=A \times B$. We then manipulate the feature vector to seem like proxy sequential data of sequence length $A$ and vector dimension $B$. The trick, therefore, can help us utilize the episodic nature of RNNs to gauge sequential correlations in our data.
    
    \item \textbf{Long Short Term Memory (LSTM) Network}: We use a three-layer LSTM network with a final softmax classification layer, similar to the one used for the RNN model. We make use of LSTMs to mitigate the widely known vanishing gradient problem \cite{vanishing_grad} of simple RNNs. We follow the same heuristical procedure to make our feature vectors suitable for training a sequential LSTM network. We believe that the LSTM should further help capture sequential dependencies inherent in our feature vectors.
\end{itemize}
\subsection{Performance evaluation}
The performance of the classification, as well as regression models, were evaluated on the test dataset that was kept separate from the training and validation process (see Figure \ref{fig:train-test}). Accuracy and mean absolute error (MAE) were used as the performance evaluation metric for the classification and regression models, respectively. The accuracy is defined as the ratio of the number of correctly predicted instances and the number of instances tested. MAE is defined as an average of absolute differences between the actual and predicted values. The accuracy could be biased in cases where the number of instances for each class are unequal. However, as we had applied SMOTE to oversample the instances of minority classes and make the number of instances belonging to each class equal, accuracy in our case is an unbiased measure. 

\begin{table*}[!htp]
\scriptsize
\caption{Percentage accuracies (the higher, the better) obtained by different ML and DL algorithms for \textbf{gender classification}. Arrangement-wise, Combined-Free-CNN (93.02\%) outperformed the rest. Device-wise, Combined (93.02\%), Phone (88.37\%), Desktop (86.04\%), and Tablet (83.33\%) closely followed each other in that order.}
\centering
\resizebox{\textwidth}{!}{
\begin{tabular}{c|c|cccccc|cccc}
\toprule
\textbf{Device} & \textbf{Setting} & \textbf{Naive Bayes} & \textbf{SVM} & \textbf{Decision Trees} & \textbf{AdaBoost} & \textbf{MLP} & \textbf{XGBoost} & \textbf{RNN} & \textbf{LSTM} & \textbf{FC} & \textbf{CNN}\\
\midrule
\textbf{Desktop} & \textbf{Free} &  72.09  & 81.39 &   76.74  &  81.39  &  \textbf{83.72}  &  \textbf{83.72}   & 77.50                               & 72.50                                & 72.09                  & \textbf{86.04}                            \\
 & \textbf{Fixed} &    72.09   &  \textbf{86.04} &  79.06   &  81.39  & 74.41   &  79.06   &                              77.50  &                                77.50 &                  62.50 & \textbf{82.50}                            \\
\midrule 

\textbf{Phone}  & \textbf{Free} & 53.48 & \textbf{83.72} &  67.44  &  81.39  &  76.74  &  81.39  & \textbf{80.00}                               & 75.00                                & 67.44                  & 79.07                            \\
 & \textbf{Fixed} &  55.81   &  \textbf{76.74}  &   74.41  &  74.41  & 72.09   &  72.09    &                            75.00    &                                85.00 &                  62.79 &  \textbf{88.37}                           \\
\midrule 

\textbf{Tablet}  & \textbf{Free} &  60.46  & \textbf{79.06} & 76.74 &  76.74  &  76.74 &  \textbf{79.06}  & \textbf{83.33}                               & 72.50                                & 69.76                  & 79.06                            \\
 & \textbf{Fixed} &   67.44   &   \textbf{72.09}  &   67.44  &  \textbf{72.09}  &   67.44  &  67.44   &                             \textbf{82.5}   &                             75.00    &                  65.11 & 79.07                            \\
\midrule 

\textbf{Combined}  & \textbf{Free} &  67.44   &\textbf{83.72} &  79.06  &  79.06  &  76.74  &   81.39   & 80.00                               & 77.50                                & 74.42                           & \textbf{93.02}                   \\

 & \textbf{Fixed} &  67.44  &  79.06  &   74.41  &   \textbf{81.39}  &  74.41  &   72.09   &                            77.50    &                               62.50  &                  67.44 & \textbf{83.72}                            \\
\bottomrule
\end{tabular}}
\label{tab:gender}
\end{table*}

\begin{table*}[!htp]
\scriptsize
\caption{Percentage accuracies (the higher, the better) obtained by different ML and DL algorithms for \textbf{major/minor classification}. Arrangement-wise, Combined-Fixed-CNN (87.80\%) outperformed the rest. Device-wise, Combined (87.80\%), Desktop (85.37\%), Tablet (85.0\%), and Phone (82.92\%) closely followed each other in that order. The results align with the with common intuition that CS majors may be more comfortable and fluent on Desktop and Tablet keypads compared to Phone than non-CS majors.}
\centering
\resizebox{\textwidth}{!}{%
\begin{tabular}{c|c|cccccc|cccc}
\toprule
\textbf{Device} & \textbf{Setting} & \textbf{Naive Bayes} & \textbf{SVM} & \textbf{Decision Trees} & \textbf{AdaBoost} & \textbf{MLP} & \textbf{XGBoost} & \textbf{RNN} & \textbf{LSTM} & \textbf{FC} & \textbf{CNN}\\
\midrule
\textbf{Desktop} & \textbf{Free} & 68.29  & \textbf{78.04} & 73.17 &  73.17  &  73.17  &   73.17      & \textbf{80.00}                               & 75.00                                & 70.73                  & 78.04                            \\
 & \textbf{Fixed} &  75.60  &  70.73  &  70.73   &  75.60  &  60.97  &  \textbf{78.04}   &                               67.50 &                               70.00  &                  56.09 &   \textbf{85.37}                          \\
 \midrule 

\textbf{Phone}  & \textbf{Free} & 60.97 & 51.21 &   \textbf{70.73}  &  65.85  &  53.65  &    53.65    & 75.00                               & 77.50                               & 68.29                 & \textbf{82.92}                            \\
 & \textbf{Fixed} &   63.41  &  60.97  &   \textbf{68.29}  &  58.53  &  58.53  &  53.65   &                               72.50 &                                77.50 &                  63.41 & \textbf{78.04}                            \\
 \midrule 

\textbf{Tablet}  & \textbf{Free} & 63.41 & 53.65 &  68.29  &   \textbf{73.17}  &  53.65  &   58.53   & \textbf{83.33}                               & 82.50                                & 68.29                  & 82.92                            \\
 & \textbf{Fixed} &   \textbf{75.60}  &  56.09  &  68.29   &   73.17   &  56.09  &  73.17   &                            72.50    &                                \textbf{85.00} &                  63.41 & 78.04                            \\
 \midrule 

\textbf{Combined} & \textbf{Free} & 65.85 & \textbf{75.60}  &  73.17  &  68.29  &  65.85 &  68.29   & 85.00                               & 80.00                                & 73.17                           & \textbf{85.37}                   \\
 & \textbf{Fixed} &  70.73  &  \textbf{73.17}  &   63.41   &  68.29  &  53.65  &  60.97   &                            82.50    &                                72.50 &                  65.85 & \textbf{87.80}                            \\
\bottomrule
\end{tabular}}
\label{tab:major_minor}
\end{table*}

\begin{table*}[!htp]
\scriptsize
\caption{Percentage accuracies (the higher, the better) obtained by different ML and DL algorithms for \textbf{typing style classification}. Arrangement-wise, Combined-Free-SVM (96.15\%) was closely followed by Combined-Fixed-SVM (94.23\%) and outperformed the rest. Device-wise, Combined (96.15\%), Phone (96.15\%), Tablet (95.55\%), and Desktop (93.18\%) closely followed each other in that order. The results do not fall beyond our expectations as we hypothesized that the typing patterns of individuals who look, occasionally look, and never look at the keypad to be very different, in general.}
\centering
\resizebox{\textwidth}{!}{%
\begin{tabular}{c|c|cccccc|cccc}
\toprule
\textbf{Device} & \textbf{Setting} & \textbf{Naive Bayes} & \textbf{SVM} & \textbf{Decision Trees} & \textbf{AdaBoost} & \textbf{MLP} & \textbf{XGBoost} & \textbf{RNN} & \textbf{LSTM} & \textbf{FC} & \textbf{CNN}\\
\midrule
\textbf{Desktop}  & \textbf{Free} &  77.27  & \textbf{93.18} &  76.92  &  90.38  &  86.53  &  81.81  & 80.00                              & 83.33                                & 82.85                  & \textbf{91.42}                            \\
 & \textbf{Fixed} &   76.92   &  \textbf{90.38} &  86.53   &  \textbf{90.38}  &  \textbf{90.38}  &   88.46  &                            50.00    &                              48.00   &                  \textbf{82.14} & 66.07                            \\
 \midrule 

\textbf{Phone} & \textbf{Free} & 78.84 & \textbf{88.63} &  82.69  &   86.36  &  86.53  &  86.36  & 83.33                               & 83.33                                & 80.70                  & \textbf{85.71}                            \\
 & \textbf{Fixed} &   86.53   &   \textbf{96.15}  &   80.76  &  84.61   & \textbf{96.15}   &  86.53   &                             50.00   &                               42.00  &                  \textbf{91.22} & 49.12                            \\
 \midrule 

\textbf{Tablet} & \textbf{Free} & 65.38 & \textbf{95.55} &  82.22  &  82.69  &  78.84  &   80.00   & 86.67                               & 83.33                                & \textbf{90.47}                  & 82.85                            \\
 & \textbf{Fixed} &   78.84 &  \textbf{90.38}  &   78.84  &  82.69  &  88.46  &   88.46   &                               56.00 &                                44.00 &                  \textbf{78.57} & 57.14                            \\
 \midrule 

\textbf{Combined} & \textbf{Free}  & 76.92 & \textbf{96.15} &  82.69  &  88.46  &  92.30  &  94.23   & 83.33                               & 80.00                                & 84.21                           & \textbf{88.57}                   \\
 & \textbf{Fixed} &  86.53  &  \textbf{94.23}  &  82.69   &  90.38  &  90.38  &   90.38  &                               70.00 &                                56.00 &                  \textbf{89.47} & 64.91                            \\

\bottomrule
\end{tabular}}
\label{tab:typing}
\end{table*}

\begin{table*}[!htp]
\scriptsize
\caption{MAE (the lower, the better) for \textbf{age and height estimation}. Arrangement-wise, Phone-Free-FC (1.77 years) and Phone-Fixed-KNN (2.65 inches) were the best performers. Device-wise, Phone (1.77 years), Desktop (2.04 years), Tablet (2.09 years), Combined (2.11 years) closely followed each other in that order. Similarly, Phone (2.65 inches), Combined (2.67 years), Tablet (2.74 inches), Desktop (2.82 inches) closely followed each other in that order. Interesting observation here is that ML algorithms have outclassed the DL algorithms. }
\centering
\begin{tabular}{c||c||ccc|cccc||ccc|cccc}
\toprule
& & \multicolumn{7}{c||}{\textbf{Age}} & \multicolumn{7}{c}{\textbf{Height}} \\ 
\midrule
\textbf{Device} & \textbf{Free/Fixed} & \textbf{SVM}  & \textbf{KNN} & \textbf{XGBoost} & \textbf{RNN} & \textbf{LSTM} & \textbf{FC} & \textbf{CNN} & \textbf{SVM}  & \textbf{KNN} & \textbf{XGBoost} & \textbf{RNN} & \textbf{LSTM} & \textbf{FC} & \textbf{CNN}\\
\midrule
\textbf{Desktop} & \textbf{Free}  & 2.37   &  2.38 &  \textbf{2.26} & 5.53                              & \textbf{2.24}                               & 2.26                  & 3.78                &  2.97  &  3.02  & \textbf{2.84} & 8.67                       & 10.70                               & 7.33                  & \textbf{7.21}                \\
 & \textbf{Fixed} & 2.43  &  2.54   &               \textbf{2.27}               & 5.24 &  \textbf{2.04}  &  2.92  & 4.97                &  2.92  &  3.20   &       \textbf{2.82}       &               9.54 &    10.66  &  8.63  & \textbf{7.24} \\
 \midrule 

\textbf{Phone} & \textbf{Free} & 2.46  &  \textbf{2.41}  & 2.59 & 7.11                               & 2.03                                & \textbf{1.77}                  & 6.10                          &  2.94   &  3.04   & \textbf{2.70} & 10.43                               & 10.39                                & \textbf{4.75}                  & 7.20                            \\
& \textbf{Fixed} & 2.38 &   \textbf{2.36}   &           2.42            &      8.41 &    2.48 &   \textbf{2.36} &  5.44              &  2.87   &   \textbf{2.65}  &       2.92        &              10.55 &  11.10    &  \textbf{5.72}  & 7.20                \\
\midrule 

\textbf{Tablet} & \textbf{Free} &  2.42  &    2.47 & \textbf{2.38} & 6.19                               & 2.45                                & \textbf{2.39}                  & 5.02                           &  \textbf{2.85}  &   3.18 & 3.23 & 8.75                               & 9.57                                & \textbf{4.83}                  & 7.22                          \\
& \textbf{Fixed} &  2.43  &  2.49   &     \textbf{2.34}           &             9.41 &     2.73 &  \textbf{2.09}  & 5.20                &  \textbf{2.74}  &   2.95  &     3.02    &                    8.42 &  9.95    &  \textbf{5.74}  & 7.20                \\
\midrule 

\textbf{Combined} & \textbf{Free} &  2.37   &  2.40  & \textbf{2.21} & 5.61                               & \textbf{2.23}                              & 2.84                           & 5.41                  &  \textbf{2.93}   &   2.99 &  3.23 & 8.52                               & 9.16                              & \textbf{7.06}                           & 7.20                  \\
& \textbf{Fixed} &   2.32   &  2.34  &       \textbf{2.27}          &          9.17   &   \textbf{2.11}   &  3.63  & 4.33                &  3.09  &  3.01   &  \textbf{2.67}     &                     7.79  &     10.61 &  11.57  & \textbf{7.20}                \\
\bottomrule
\end{tabular}
\label{tab:AgeHeight}
\end{table*}

\begin{table*}[htp]
\caption{Qualitative comparison with previous works that attempted to infer the soft biometrics that we have considered. kFCV means k-Fold Cross-Validation, while HOCV means Hold the test set Out Cross-Validation in this study (see Figure \ref{fig:train-test}). We achieved almost perfect Accuracy and MAE between 1-2 for both age and height under kFCV. We are not reporting kFCV results as it is a less realistic evaluation setup than HOCV, especially for the application scenarios listed in this paper. \vspace{0.02in}
}
\small
\centering
\label{tab:prevworkcomparison}
\begin{tabular}{llllllc}
\toprule
\textbf{Ref.}                                           & \textbf{Users} & \textbf{\begin{tabular}[c]{@{}l@{}}Free/Fixed\end{tabular}} & \textbf{\begin{tabular}[c]{@{}l@{}}Class\end{tabular}} & \textbf{\begin{tabular}[c]{@{}l@{}}Desktop/Phone\end{tabular}} & \textbf{\begin{tabular}[c]{@{}l@{}}kFCV/HOCV\end{tabular}} & \textbf{\begin{tabular}[c]{@{}l@{}}Accuracy/MAE\end{tabular}}                                                             \\ 
\midrule

\cite{ImprovedKDAuthWithGenderPred2012}                   & 133           & Fixed                                                          & Gender                                                         & Desktop                                                           & kFCV                                                          & 91.63                                                                       \\ 
\midrule 

\cite{FirstGenderFromKeystroke2011}                   & 133           & Fixed                                                          & Gender                                                         & Desktop                                                           & kFCV                                                          & 97.50                                                                       \\ 
\midrule 
\cite{ChildVsAdultKeystroke}                          & 100           & Fixed                                                          & Age                                                            & Desktop                                                           & kFCV                                     & 91.20                                                                       \\ 
\midrule 

\cite{AgeGenderKDMouseMove2017}                       & 1519          & Both                                                           & Both                                                           & Desktop                                                           & kFCV                                     & 73.00                                                                         \\ 
\midrule 

\cite{plank2018predicting}                            & 144           & Free                                                           & \begin{tabular}[c]{@{}l@{}}Age \\ Gender\end{tabular}                                                           & Desktop                                                           & kFCV                                     & \begin{tabular}[c]{@{}l@{}}63.50 \\ 73.25\end{tabular} \\ 
\midrule 

\cite{Gender2018FromKeystroke}                        & 75            & Free                                                           & Gender                                                         & Desktop                                                           & kFCV                                     & 95.60                                                                       \\ 
\midrule 

\cite{CombiningKDandStylometryGenderPredFromChat2019} & 45            & Free                                                           & Gender                                                         & Desktop                                                           & kFCV                                     & 72.00                            \\  \midrule
\cite{GenderFromChat2020}                             & 60            & Free                                                           & Gender                                                         & Desktop                                                           & kFCV                                     & 98.30                                                                       \\ 
\midrule
\cite{AgeGendernumFingers2014}                        & 132           & Fixed                                                          & \begin{tabular}[c]{@{}l@{}}Age \\ Gender\end{tabular}                                                           & Phone                                                             & HOCV                                                          & \begin{tabular}[c]{@{}l@{}}60.30 \\ 75.20\end{tabular}                        \\ 
\midrule 
\cite{GenderAgeKeystrokeHamdedness2014FreeFixedBoth}  & 110           & Both                                                           & \begin{tabular}[c]{@{}l@{}}Age \\ Gender\end{tabular}                                                           & Desktop                                                           & HOCV                                                          & \begin{tabular}[c]{@{}l@{}}78.00 \\ 86.00\end{tabular}                      \\ 
\midrule 

\cite{AgeGenderHandMobile2016}                        & 150           & Fixed                                                          & \begin{tabular}[c]{@{}l@{}}Age \\ Gender\end{tabular}                                                           & Phone                                                             & HOCV                                                          & \begin{tabular}[c]{@{}l@{}}82.80 \\ 87.70\end{tabular}                        \\ 
\midrule 

\cite{CombineKDandSignatueForGenderPrediction}        & 100           & Both                                                           & Gender                                                         & Desktop                                                           & HOCV                                                          & 71.30                                                                       \\ 
\midrule 

\textbf{This work}                                              & 117           & Free                                                           & \begin{tabular}[c]{@{}l@{}}Gender \\ Major \\ Style \\ Age \\ Height \end{tabular}                                                          & \begin{tabular}[c]{@{}l@{}} The best of \\ Desktop, \\ Phone, Tablet, \\ and Combined \end{tabular}                                                             & HOCV                                                       & \begin{tabular}[c]{@{}l@{}} 93.02\\ 85.37 \\ 96.15\ \\ 1.77 \\ 2.70 \end{tabular} \\ \midrule

\textbf{This work}                                              & 117           & Fixed                                                           & \begin{tabular}[c]{@{}l@{}}Gender \\ Major \\ Style \\ Age \\ Height \end{tabular}                                                          & \begin{tabular}[c]{@{}l@{}} The best of \\ Desktop, \\ Phone, Tablet, \\ and Combined \end{tabular}                                                             & HOCV                                                       & \begin{tabular}[c]{@{}l@{}} 88.37 \\ 87.80 \\ 96.15 \ \\ 2.04 \\ 2.65 \end{tabular}    

\\ \bottomrule
\end{tabular}
\end{table*}

\section{Results}
\label{sec:Results}
\subsection{Classification results}
The following subsections discuss the results obtained by different ML and DL based classification models used in this study:
\subsubsection{Gender classification}
The gender classification accuracies are presented in Table \ref{tab:gender}. In terms of devices, the combined case achieved the best results (93.02\%) followed by Phone (88.37\%), Desktop (86.04\%), and Tablet (83.33\%). Free-text (93.02\%) yielded better results than the Fixed-text (88.37\%), overall. Classifier-wise, CNN (93.02\%), SVM (86.04\%), MLP/XGBoost (83.72\%), and RNN (83.33\%) outperformed the rest. 
\subsubsection{Major/Minor classification} 
The accuracies for the major/minor classification task can be found in Table \ref{tab:major_minor}. In terms of devices, the combined-device setting achieved the best results (87.8\%) followed by Desktop (85.37\%), Tablet (85\%), and Phone (82.92\%). Overall, Fixed-text (87.8\%) yielded slightly better results than Free-text (85.37\%). The top-performing classifiers were CNN (87.8\%), LSTM (85\%), RNN (83.33\%), SVM (78.04\%) and XGBoost (78.04\%) followed by the rest.
\subsubsection{Typing style classification} 
The accuracies for the typing style classification task can be found in Table \ref{tab:typing}. In terms of devices, the combined-device setting and Phone achieved the best results (96.15\%) followed by Tablet (95.55\%), and Desktop (93.18\%). Overall, both Fixed-text and Free-text yielded the same best results (96.15\%). The top-performing classifiers were SVM (96.15\%), MLP (96.15\%), CNN (91.42\%), FC (91.22\%) and AdaBoost (90.38\%) followed by the rest. 
\subsection{Regression results}
The following subsections discuss the results obtained by different ML and DL based regression models used in this study:
\subsubsection{Age estimation}
The collated results for both ML and DL models for the task of age prediction can be found in Table \ref{tab:AgeHeight}. In terms of devices, the phone-only setting achieved the best results (1.77) followed by desktop (2.04), tablet (2.09), and combined (2.11). Free-text (1.77) yielded better results than the Fixed-text (2.04), overall. Regressor-wise, FC (1.77), LSTM (2.04), and XGBoost (2.21) outperformed the rest. 
\subsubsection{Height estimation} 
The results for both the ML and DL models for the height prediction problem can be found in Table \ref{tab:AgeHeight}. In terms of devices, the phone-only setting achieved the best results (2.65) followed by combined (2.67), tablet (2.74), and desktop (2.82). In contrast to age regression, Fixed-text (2.65) yielded better results than the Free-text (2.70), overall. Regressor-wise, KNN (2.65), XGBoost (2.67), and SVM (2.74) outperformed the rest. For the height prediction problem, ML regressors clearly outperformed DL regressors.

\section{Discussion}
\label{sec:Discussion}
\subsection{Limitations} 
As mentioned earlier, one of the major limitations of studying the inference of soft biometrics is a quality dataset. Although every participant provided about thirty thousand keystrokes, the number of subjects is limited in the dataset, which makes the training, validation, and testing a bit difficult. In particular, we used the data collected from 70\% users (i.e., 82 users) for training and cross-fold validation, while the data collected from the rest and the data collected from the remaining 30\% (i.e., 35 users) used for testing. Another limitation of the dataset is that the samples for recorded soft biometrics are severely imbalanced in some cases (see Table \ref{tab:samplestat}). For example, of the total 117 participants, 105 are Asian, and 114 identified themselves as right-handed. 

Although we expect that the performance of the proposed approaches would scale to a larger dataset, it is difficult to claim that it would. Nonetheless, the results are comparable or better than the existing mechanisms of inferring soft biometrics from keystrokes (see Table \ref{tab:prevworkcomparison}).

\subsection{Ethical implications} While in the introduction section, we have listed positive application scenarios, people with malicious intent can use the research presented in this work for destructive purposes. We, however, believe that the misuse can be prevented by developing existing as well as new public policies \cite{plank2018predicting}. 

\section{Conclusion and future work}
\label{sec:ConclusionAndFutureWork}
We conclude that soft biometrics such as gender, typing style, major/minor, age, and height can be inferred from typing patterns of individuals with reasonably good accuracy. The free-text analysis showed more promise compared to the fixed-text environment except for the major/minor prediction. DL methods outclassed ML methods overall except for the height estimation task. The Phone-Fixed configuration achieved the highest gender recognition accuracy (88.37\%), while the combination of data collected from all three devices helped better the results (93.02\%). The Desktop-Fixed setup outclassed the rest of the individual device setups achieving 85.37\% accuracy in the major/minor classification, while the combined experimental setup reached 87.80\%. The Phone-Fixed configuration attained the highest accuracy in typing style classification, and the combination of data from multiple devices did not help in this case. The Phone-Free setting predicted the age with an MAE of 1.77 years, while the Phone-Fixed setup estimated the height with an MAE of 2.65 inches. 

We would like to test the proposed approaches on multiple datasets, especially on a dataset that consists of participants of more diverse backgrounds and demographics. Besides, we would also like to investigate other modalities available in the SU-AIS BB-MAS dataset as well as the fusion of all the modalities for better estimation and prediction of the soft biometrics. In the end, based on our observation that the deep learning-based models outsmarted traditional ML algorithms, we would like to leverage more deep learning architectures for their effectiveness in the soft biometric prediction task. 

\bibliographystyle{ACM-Reference-Format}
\bibliography{references_full}
\end{document}